\documentclass[preprint,12pt,number]{elsarticle}
\usepackage{amssymb}
\usepackage{comment}
\usepackage{multirow}
\usepackage{tabularx}
\usepackage{amsmath}
\usepackage{enumitem}
\usepackage[colorlinks=true,linkcolor=blue]{hyperref}
\usepackage{xurl}
\usepackage{xcolor}
\usepackage{wasysym}
\usepackage{textcomp}
\usepackage[T1]{fontenc}

\journal{To be determined}
\begin{document}
\begin{frontmatter}

\title{AI and the Law: Evaluating ChatGPT's Performance in Legal Classification}

\author[inst1]{Paweł Weichbroth*}
\affiliation[inst1]{organization={Gdansk University of Technology, Faculty of Electronics, Telecommunications and Informatics, Department of Software Engineering},
            city={Gdansk},
            country={Poland}, \newline *Corresponding Author: pawel.weichbroth@pg.edu.pl}

\begin{abstract}
The use of ChatGPT to analyze and classify evidence in criminal proceedings has been a topic of ongoing discussion. However, to the best of our knowledge, this issue has not been studied in the context of the Polish language. This study addresses this research gap by evaluating the effectiveness of ChatGPT in classifying legal cases under the Polish Penal Code. The results show excellent binary classification accuracy, with all positive and negative cases correctly categorized. In addition, a qualitative evaluation confirms that the legal basis provided for each case, along with the relevant legal content, was appropriate. The results obtained suggest that ChatGPT can effectively analyze and classify evidence while applying the appropriate legal rules. In conclusion, ChatGPT has the potential to assist interested parties in the analysis of evidence and serve as a valuable legal resource for individuals with less experience or knowledge in this area.
\end{abstract}


\end{frontmatter}

\section{Introduction}
\label{sec:introduction}
The law is a fundamental part of our daily lives \cite{galligan2006law}, and it is almost inevitable that everyone will interact with the justice system at some point. Legal professionals, such as attorneys, prosecutors, and judges, apply the law according to their specific roles, including advocating for clients, investigating criminal activity, and adjudicating cases. 
Meanwhile, the other parties involved, especially those without prior expertise and relevant knowledge, may find themselves confused in legal proceedings as defendants or plaintiffs, whether they have been accused of violating the law or their rights have been violated or their rights have been violated by others. Although the principle of \textit{ignorantia iuris nocet} (``ignorance of the law is harmful'' \cite{szylkowska2019ignorantia}) does not always apply here, interpreting and analyzing legal regulations remains a daunting task for most, or even in some cases barely feasible.

The advent of artificial intelligence (AI) has been a game changer in many fields \cite{pondel2024ai, yuan2024artificial, rampavsek2025evolving}, including law \cite{ashley2019brief}. A recent report, published by Thomson Reuters in 2024 \cite{TR-2024}, highlights how AI is reshaping the legal profession by automating routine tasks and increasing lawyer productivity. AI-powered tools streamline document review, legal research, and contract analysis, allowing lawyers to focus on higher-value work. Note that that AI could also save lawyers four hours per week while generating an additional \$100,000 in billable time per lawyer per year.

On the other hand, AI-based tools can help people with little or no legal education understand legal language by simplifying complex legal terminology into plain, easy-to-understand text \cite{weber2024legalwriter}. Such tools can break down statutes \cite{surden2021machine}, contracts \cite{autto2024contracts}, and court decisions \cite{habernal2024mining} into concise summaries so that key implications can be grasped. In addition, AI can provide contextual explanations and real-world examples to illustrate how legal principles apply to specific situations \cite{prakken2015law}. By making legal information more accessible, AI empowers individuals to make informed decisions before seeking professional legal assistance \cite{lorek2024ai}.

One such AI-based tool is \href{https://chatgpt.com/}{ChatGPT}. Developed by OpenAI \cite{lozic2023fluent}, an AI research and deployment company \cite{hang2024large}, it is a state-of-the-art conversational agent built upon the Generative Pre-trained Transformer (GPT) architecture \cite{yenduri2024gpt}. Its development represents a significant milestone in the field of AI \cite{wu2023brief}, particularly in natural language processing (NLP) \cite{Belcic2024}. The evolution of ChatGPT can be traced back to earlier iterations of GPT models \cite{banik2024systematic}, with each version introducing advancements in language understanding and generation \cite{salloum2024coming}. The release of ChatGPT has garnered considerable attention \cite{khan2024global}, leading to discussions about its applications, limitations , and potential future developments \cite{dale2021gpt, sohail2023future, kozachek2023investigating}. 

With these opportunities and challenges in mind, it is important to carefully evaluate the role of ChatGPT in the legal field. Understanding its strengths and limitations can help determine its potential use in a specific context. That being said, this study aims to explore the potential and limitations of ChatGPT for the Polish language by analyzing its effectiveness in classifying legal cases in terms of the Penal Code. By conducting a qualitative and quantitative analysis of ChatGPT's performance in legal tasks, this study seeks to provide insights into the extent to which AI is able to correctly infer, classify and explain evidence written in the form of short text notes. 

The rest of the paper is structured as follows. 
Section~\ref{sec:related-work} discusses related work.
Section~\ref{sec:methodology} describes the research methodology.
Section~\ref{sec:Results} presents the results evaluation.
Section~\ref{sec:discussion} outlines the further discussion on the research, in particular idicating its contribution and limitations.
Section~\ref{sec:conclusions} concludes the paper.

\section{Related Work}
\label{sec:related-work}
There is no doubt that the emergence of advanced tools such as ChatGPT has brought new opportunities to enhance law enforcement \cite{pandey2024exploring}, risk management \cite{halford2024using}, improve efficiency \cite{raj2023analyzing}, and expand operational capacity \cite{chen2024antecedents}. The embedded AI-driven capabilities can be leveraged across multiple tasks that have been explored in different scenarios and tailored settings.

Perlman \cite{perlman2023implications} demonstrated the use cases based on prompting ChatGPT with questions to test its capabilities in document generation, legal analysis, information and research. The author stated that the obtained responses ``were imperfect and at times problematic", while the generated documents were incomplete. Nevertheless, attorneys may be willing to prepare initial drafts of complex legal instruments, easily adopting the firm's style and incorporating its substantive knowledge. Interestingly, one of the author's other closing observations was that ``AI will not eliminate the need for lawyers, but it does portend the end of lawyering as we know it". We also agree that the AI-driven revolution in legal practice is undeniable, and resistance seems futile.

Armstrong \cite{armstrong2023s} found that ChatGPT-4 was able, although sometimes imperfectly, to engage in legal reasoning about law and factual data, answer questions about legal opinions or documents, and even analyze, draft and summarize legal cases, write motions, draft patents, and write reports. However, in some cases ChatGPT performed particularly poorly on this set of prompts, occasionally producing unsatisfactory results with some serious errors. To conclude, since ChatGPT presented fiction as fact (and with confidence), legal professionals should not rely on ChatGPT responses, and thus it is not a substitute for careful reading and case analysis. 

At the University of Minnesota Law School, Choi et al. \cite{choi2021chatgpt} tested ChatGPT to generate answers to four real exams, including Constitutional Law: Federalism and Separation of Powers, Employee Benefits, Taxation, and Torts. In all four courses, over 95 multiple-choice questions and 12 essay questions, its performance was on average at the level of a C+ student, achieving a low but passing grade. In general, ChatGPT performed better on the essay sections of the exams than on the multiple choice questions. Note that if this performance were consistent throughout law school, its grades would be sufficient for a student to graduate.

Jan and others \cite{tan2023chatgpt} asked a stirring question: ChatGPT as an artificial lawyer? In general, the results obtained were not accurate enough to deliver legal information directly to inexperts. ChatGPT sometimes generates false information with high confidence, especially about laws and cases. In this regard, it was less willing to revise answers, even when challenged multiple times. However, ChatGPT was able to provide a seamless interactive experience with a minimal learning curve, allowing users to describe their legal issues in fragmented language and clarify or refine details as the conversation progressed. 

Apparently, there is no consensus among researchers about the effectiveness of ChatGPT in analyzing legal documents. Some studies highlight its potential for drafting legal documents and assisting with legal reasoning, while acknowledging its frequent inaccuracies and incomplete results. Others emphasize its ability to pass legal assessments at a basic level, suggesting limited but still useful capabilities. 

However, concerns persist about its tendency to generate incorrect legal information with high confidence, making it unreliable for professionals who require precise legal analysis. While some see ChatGPT as a game-changing tool in legal practice, others caution against overreliance due to its apparent limitations. These differing perspectives suggest that while such a tool holds promise, its role in legal practice remains a subject of ongoing debate.

In light of the above, we are pleased to join this lively discussion, and in our study we aim to investigate the effectiveness of ChatGPT by testing its capabilities on a self-developed Polish language dataset in the classification of evidence under the Polish Penal Code.

\section{Methodology}
\label{sec:methodology}

The research problem concerns the analysis and classification of evidence collected in the form of a text note, performed by ChatGTPT, in the course of proceedings conducted by law enforcement agencies and the judgment of whether it is a criminal offense or not. 
More specifically, classification refers to the application by ChatGPT of the relevant provisions of the Polish Penal Code, which indicate the positive (P) or negative (N) qualification of the act, taking into account the effects of the actions of the suspect(s).

Note that, law enforcement agencies in Poland refer to: Police, Central Anti-Corruption Bureau (CBA), Internal Security Agency, Polish Border Guard, Customs Service, and Military Police.

The collection of input data was prepared as follows. Based on the information available on the Internet, a collection of 268 text notes was compiled, including both positive and negative cases. Thus, the cardinality of both classes was equally represented in the dataset.

Each case was written by hand, and consists of three up to five sentences, describing a unique event, embedded in a place and time. The positive case exhibits an unambiguous indication of the person suspected of committing exactly one criminal act, as defined in the Penal Code or in another currently applicable legal document such as a law or regulation. 

The negative case, on the other hand, describes a suspicious situation in which there are some premises of a possible violation of the law, but which objectively do not constitute grounds for an illegal act.

A set of positive cases was manually labeled by a professional lawyer who, after analyzing the case, classified each case by precisely specifying the type of illegal act. There were no predefined labels for use. In practice, this meant searching and analyzing the applicable laws and selecting and assigning the relevant paragraph to a particular case. 

In total, 15 different types of illegal acts have been identified and specified. Table~\ref{tab:crimes-list} shows the original meaning in Polish, the translation into English and the legal basis. 

Negative cases, on the other hand, refer to situations where individuals came close to breaking the law but ultimately did not commit a violation, making it invalid to classify their actions as positive. For instance, consider the following case. 
On 29 November 2024, at 9:30 a.m., a Honda CR-V (registration number QWE99098) was stopped for inspection on Narutowicza Street 11 in Gdansk. The driver, Milosz Sz., underwent a breathalyzer test, which registered 0.005 mg/L (approximately 0.0105\textperthousand). It can be concluded that this level is within the legally acceptable limit according to the Polish regulations. Therefore, the classification of such conduct is negative.

\section{Results}
\label{sec:Results}
The evaluation of the results was carried out on the basis of a confusion matrix which is a fundamental evaluation tool used in classification tasks to measure the performance of a model by comparing predicted labels with actual labels. 
Since our study concerns binary classification, the confusion matrix, written as a table of size 2×2, with the following components, was used. Table~\ref{tab:confusion-matrix} shows the details in this extent. 

\begin{table}[h]
\caption{Confusion matrix.}
\label{tab:confusion-matrix}
\centering
\begin{tabular}{|l|l|l|}
\hline
\textbf{Actual }\textbackslash \textbf{Predicted} & \textbf{Positive} (1) & \textbf{Negative} (0) \\ \hline
\textbf{Positive}   (1)  & True   Positive (TP)  & False   Negative (FN) \\ \hline
\textbf{Negative }  (0)  & False   Positive (FP) & True   Negative (TN)  \\ \hline
\end{tabular}
\end{table}

The format used in the confusion matrix is that positive or negative refers to the class values predicted by the model, and true or false refers to the accuracy of the predicted model. Therefore, the interpretation of each cell is as follows: 

\begin{itemize}
    \item True Positives (TP). Cases where the ChatGPT correctly predicted the positive class.
    \item False Negatives (FN). Cases where the ChatGPT incorrectly predicted the negative class (Type II error).
    \item False Positives (FP). Cases where the ChatGPT incorrectly predicted the positive class (Type I error).
    \item True Negatives (TN). Cases where the ChatGPT correctly predicted the negative class.
\end{itemize}

\begin{figure}
    \centering
    \includegraphics[width=0.45\linewidth]{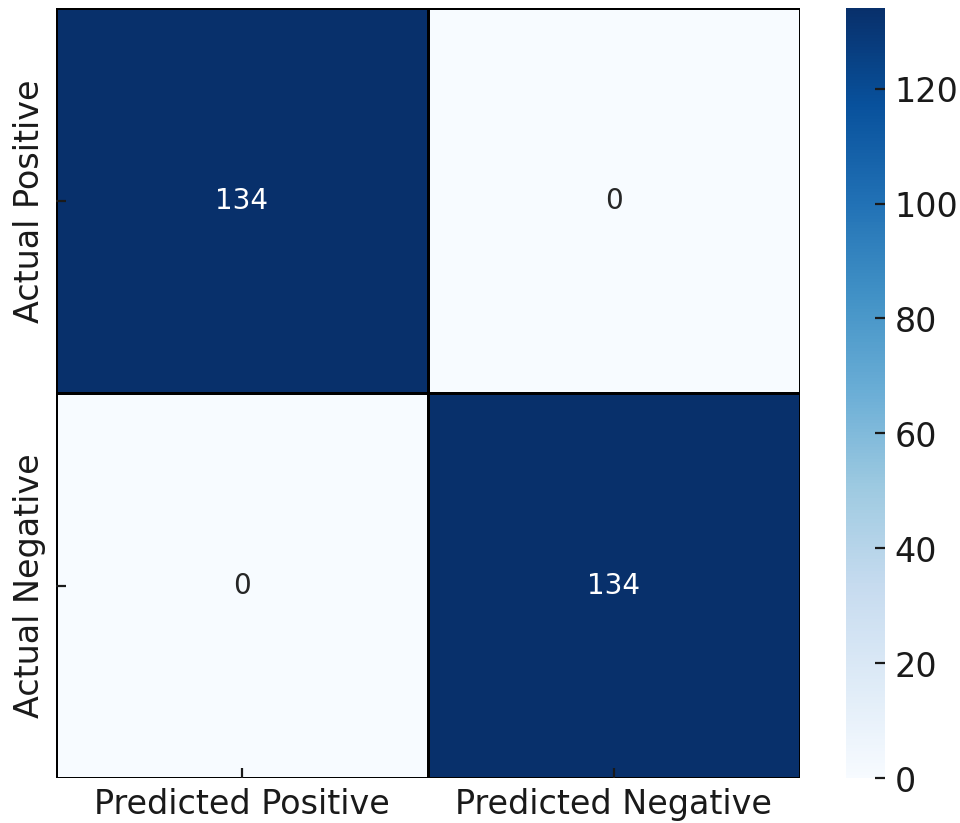}
    \caption{Confusion matrix for ChatGPT performance.}
    \label{fig:confusion-matrix}
\end{figure}

In the period from December 2024 to January 2025, ChatGPT was asked to classify a collection of positive and negative cases, typically from one to five cases in a single message. In all cases analyzed, ChatGPT output was correct, as shown in Figure~\ref{fig:confusion-matrix}. 

To evaluate ChatGPT in event classification, the following five evaluation metrics were used.

\textbf{1. Accuracy} (Overall correctness of the model):

\begin{equation} \label{eq:Accuracy}
    \text{Accuracy} = \frac{TP + TN}{TP + TN + FP + FN}
\end{equation}

Accuracy measures the proportion of correctly classified instances out of the total instances. Here, a value of 1.0 (100\%) indicates a perfect classification with no errors made by ChatGPT.

\textbf{2. Precision} (Positive Predictive Value):

\begin{equation} \label{eq:Precision}
    \text{Precision} = \frac{TP}{TP + FP}    
\end{equation}

Precision represents the proportion of correctly classified positive cases out of all predicted positive cases. In our research, a perfect precision score of 1.0 means there were no false positives (FP = 0), meaning ChatGPT never misclassified a negative case as positive.

\textbf{3. Recall} (Sensitivity or True Positive Rate):

\begin{equation}\label{eq:Recall}
    \text{Recall} = \frac{TP}{TP + FN}    
\end{equation}

Recall measures the proportion of actual positive cases that were correctly identified. A recall of 1.0 means that the ChatGPT detected all positive cases with no false negatives (FN = 0).

\textbf{4. F1 Score} (Harmonic mean of precision and recall):
\begin{equation}\label{eq:F1-Score}
    \text{F1 Score} = 2 \times \frac{\text{Precision} \times \text{Recall}}{\text{Precision} + \text{Recall}}    
\end{equation}

The F1 Score is the harmonic mean of Precision (Eq.~\ref{eq:Precision}) and Recall (Eq.~\ref{eq:Recall}), balancing both measures. Since both precision and recall are 1.0, the F1 score also reaches its maximum value, indicating perfect classification with no trade-off between FPs and FNs.

\textbf{5. Specificity} (True Negative Rate):

\begin{equation}\label{eq:Specificity}
    \text{Specificity} = \frac{TN}{TN + FP}    
\end{equation}

Specificity measures how well the model correctly identifies negative cases. A specificity of 1.0 means that all negative cases were classified correctly, with no false positives.

In summary, ChatGPT achieved perfect classification, meaning that there were no misclassifications of either positive or negative cases in a total of 238 cases analyzed. In addition, in terms of the qualitative evaluation of the explanations provided, ChatGPT provided the correct legal basis for each classification, specifying the content of each paragraph of the Polish Penal Code. 



\section{Discussion}
\label{sec:discussion}
Legal documents such as court decisions, statutes, contracts, or regulatory guidelines are often long and filled with complicated terminology. From this point of view, summarizing legal texts turns out to be one of the most valuable applications of ChatGPT for legal professionals. 

ChatGPT has proven its ability to analyze the texts and extract the most relevant points, saving lawyers considerable time and effort. It can present summaries in a structured format, highlighting key arguments, regulations, rights and precedents without omitting critical details. This is particularly useful for novice lawyers who need to quickly understand a case or legal framework without having to read through a stack of documents.

In addition, testing ChatGPT also revealed the ability to customize summaries, which seems valuable when considering a user's profession. For example, a legal professional, a corporate executive, or an individual seem to have different expectations and needs. This adaptability makes it a powerful tool for improving accessibility and efficiency in the legal field. By simplifying legal language for clients, it makes it easier for non-lawyers to understand their rights and responsibilities. 

The novelty of this paper lies in the application of ChatGPT in the analysis of the Polish language with the aim of classifying short text messages in terms of the Criminal Code. In this view, the findings concern the performance evaluation, performed in both quantitative and qualitative manner. Specifically, using a novel dataset built from the ground up and externally validated, we demonstrated ChatGPT's ability to effectively classify collected evidence by selecting valid regulations, applying relevant laws, and providing up-to-date content.

Nonetheless, our study suffers from at least two obvious limitations. The first concern is the potential lack of generalizability due to the limited number of crime types. Without testing ChatGPT on a richer dataset, there is a risk that the observed perfect performance is specific to the fifteen crime types tested and does not truly reflect its ability to classify unseen cases. 

Second, the dataset used appears to be relatively small and described in a concise manner, which may cause some researchers to question the results. However, the cases were developed based on real-world scenarios, which reduces the potential bias. Nevertheless, our dataset does not contain rare cases that could negatively affect ChatGPT's performance.

To confirm our findings, additional experiments with larger and more diverse datasets should be conducted. However, developing datasets from scratch that include real-world evidence is a considerable challenge. On the other hand, this approach raises concerns about confidentiality, in particular the privacy and security of personal data. 

\section{Conclusions}
\label{sec:conclusions}
This research evaluated ChatGPT's ability to classify legal evidence written in Polish and determine whether a crime was committed based on the Polish Penal Code. The results show that ChatGPT correctly classified all 134 positive and 134 negative cases. Considering the former type, the model demonstrated logical analysis by considering not only the input data but also relevant legal resources, ensuring their correct interpretation and application. Taking into account the later type, however, posed an even greater challenge, as they often described situations that were legally ambiguous, requiring the model to distinguish between legal and borderline illegal actions while avoiding intentional deception.

With several new features recently added to ChatGPT, the future looks even brighter and more promising. First, the level of personalization in ChatGPT has been updated by enhancing the custom instructions to respond to user's questions and requests in a more personalized manner. Second, Scheduled Tasks is a feature that allows a user to run automated prompts and proactively reach out to him/her on a scheduled basis. Third, the Projects feature provides a new way to group files and chats for personal use, making it easier to manage work that spans multiple chats.

Finally, ChatGPT can also analyze images by identifying objects, text, patterns, and overall context within them. These capabilities include reading text from images (OCR), recognizing scenes, and interpreting visual elements to provide meaningful insights. These capabilities can be used to analyze evidence in both criminal and civil cases. However, they are still limited to general recognition and cannot perform deep image processing like human experts or specialized AI models.

Undoubtedly, the story is far from over, and we can expect new features to emerge with the ability to speak and listen much like humans do. Still, many questions arise. How will ChatGPT affect our relationships with others? What are the broader implications for ChatGPT -- and AI as a whole -- in society? The shift to an era where machines can not only think but also see, hear, and speak is undeniably remarkable. However, it does not absolve us of responsibility for how we, as humans, choose to use these assets.

\section*{Data availability}
Data available on reasonable request.

\section*{Funding}
The research was supported by project “Cloud Artificial Intelligence Service Engineering (CAISE) platform to create universal and smart services for various application areas”, No. KPOD.05.10-IW.10-0005/24, as part of the European IPCEI-CIS program, financed by NRRP (National Recovery and Resilience Plan) funds.

\section*{Declaration of Interest Statement}
The author declares that he has no known competing financial interests or personal relationships that could have appeared to influence the work reported in this article.



\newpage
\section*{Appendix}

\begin{table}[h]
\centering
\caption{List of 15 crimes (positive cases) used for testing ChatGPT performance}
\label{tab:crimes-list}
\footnotesize
\begin{tabular}{|p{5cm}|p{5cm}|p{5cm}|}
\hline
\textbf{Illegal act (in Polish)}  & \textbf{Illegal act (in English) }  & \textbf{Regulation (in Polish)  }            \\ \hline
Handel narkotykami   & Drug trafficking  & Art. 62 ust. 2 Ustawy o   przeciwdziałaniu narkomanii: Art. 59 ust. 1 Ustawy o przeciwdziałaniu   narkomanii 
\\ \hline
Kradzież rozbójnicza   & Robbery   & Art. 281 Kodeksu karnego 
\\ \hline
Kradzież z włamaniem  & Burglary  & Art. 279 § 1 Kodeksu   karnego 
\\ \hline
Poświadczenie nieprawdy  & Attestation of untruth  & Art. 271 § 1 i § 3 Kodeksu karnego 
\\ \hline
Produkcja narkotyków & Drug production  & Art. 200 § 1 Kodeksu karnego
\\ \hline
Prowadzenie samochodu, motocykla, roweru w stanie   nietrzeźwości lub odurzenia  & Driving a car, motorcycle, bicycle while under the   influence of alcohol or intoxication & Art. 178a § 1 Kodeksu karnego 
\\ \hline
Przestępstwo przeciwko bezpieczeństwu w komunikacji: spowodowanie wypadku komunikacyjnego & Crimes against traffic safety: causing a traffic accident   & Art. 177 § 1 Kodeksu karnego 
\\ \hline
Przestępstwo przeciwko wiarygodności dokumentów:   fałszerstwo dokumentów  & Crimes against the credibility of documents: forgery of documents  & Art. 297 § 1 Kodeksu karnego: Art. 270 § 1 Kodeksu karnego:  
\\ \hline
Przewóz i wprowadzanie do obrotu środków odurzających  & Transportation and marketing of narcotic drugs   & Art. 55 ust. 1 oraz Art. 56 ust. 1 Ustawy o przeciwdziałaniu narkomanii
\\ \hline
Rozbój   & Robbery & Art. 280 § 1 Kodeksu karnego 
\\ \hline
Spowodowanie rozstroju zdrowia   & Causing bodily disorder  & Art. 177 § 1 Kodeksu   karnego 
\\ \hline
Udział w bójce lub pobiciu, narażenie na utratę życia lub zdrowia  & Participation in a fight or beating, exposure to loss of life or health   & Art. 158 § 1 Kodeksu karnego  
\\ \hline
Udział w bójce skutkującej ciężkim uszczerbkiem na zdrowiu  & Participation in a fight resulting in serious bodily injury  & Art. 158 § 2 Kodeksu karnego
\\ \hline
Wprowadzanie do obrotu środków odurzających  & Placing narcotic drugs on the market   & Art. 56 ust. 1 Ustawy o   przeciwdziałaniu narkomanii 
\\ \hline
wymuszenie rozbójnicze & Robbery extortion   & Art. 282 Kodeksu karnego \\ \hline
\end{tabular}
\end{table}
\end{document}